\documentclass[final]{cvpr}

\usepackage{times}
\usepackage{epsfig}
\usepackage{graphicx}
\usepackage{amsmath}
\usepackage{amssymb}
\usepackage{subfigure}
\usepackage{colortbl}
\usepackage{booktabs}
\usepackage{multirow}
\usepackage{threeparttable}
\usepackage[font=small,labelfont=bf]{caption}
\DeclareCaptionLabelFormat{andtable}{#1~#2  \&  \tablename~\thetable}
\definecolor{babyblueeyes}{rgb}{0.63, 0.79, 0.95}
\definecolor{babyblue}{rgb}{0.54, 0.81, 0.94}
\definecolor{columbiablue}{rgb}{0.61, 0.87, 1.0}

\newcommand\blfootnote[1]{%
  \begingroup
  \renewcommand\thefootnote{}\footnote{#1}%
  \addtocounter{footnote}{-1}%
  \endgroup
}

\usepackage[pagebackref=true,breaklinks=true,colorlinks,bookmarks=false]{hyperref}

\def\cvprPaperID{4240} 
\def\confYear{CVPR 2021}

\begin{document}
\title{ACTION-Net: Multipath Excitation for Action Recognition}

\author{Zhengwei Wang$^1$\hspace{20pt}Qi She$^2$\hspace{20pt}Aljosa Smolic$^1$\\
\textsuperscript{1}V-SENSE, Trinity College Dublin, Ireland\hspace{5pt} \textsuperscript{2}ByteDance AI Lab, China\\
{\tt\small\{zhengwei.wang,SMOLICA\}@tcd.ie, sheqi1991@gmail.com}
}


\maketitle

\begin{abstract}
Spatial-temporal, channel-wise, and motion patterns are three complementary and crucial types of information for video action recognition. Conventional 2D CNNs are computationally cheap but cannot catch temporal relationships; 3D CNNs can achieve good performance but are computationally intensive. In this work, we tackle this dilemma by designing a generic and effective module that can be embedded into 2D CNNs. To this end, we propose a sp\textbf{A}tio-temporal, \textbf{C}hannel and mo\textbf{T}ion excitat\textbf{ION} (ACTION) module consisting of three paths: Spatio-Temporal Excitation (STE) path, Channel Excitation (CE) path, and Motion Excitation (ME) path. The STE path employs one channel 3D convolution to characterize spatio-temporal representation. The CE path adaptively recalibrates channel-wise feature responses by explicitly modeling interdependencies between channels in terms of the temporal aspect. The ME path calculates feature-level temporal differences, which is then utilized to excite motion-sensitive channels. We equip 2D CNNs with the proposed ACTION module to form a simple yet effective ACTION-Net with very limited extra computational cost. ACTION-Net is demonstrated by consistently outperforming 2D CNN counterparts on three backbones (i.e., ResNet-50, MobileNet V2 and BNInception) employing three datasets (i.e., Something-Something V2, Jester, and EgoGesture). Codes are available at \url{https://github.com/V-Sense/ACTION-Net}.
\end{abstract}

\section{Introduction}
\blfootnote{This work is financially supported by Science Foundation Ireland (SFI) under the Grant Number 15/RP/2776. We gratefully acknowledge the support of NVIDIA Corporation with the donation of the NVIDIA DGX station used for this research.}
\begin{figure*}[!htbp]
    \centering
    \includegraphics[width=1.\textwidth]{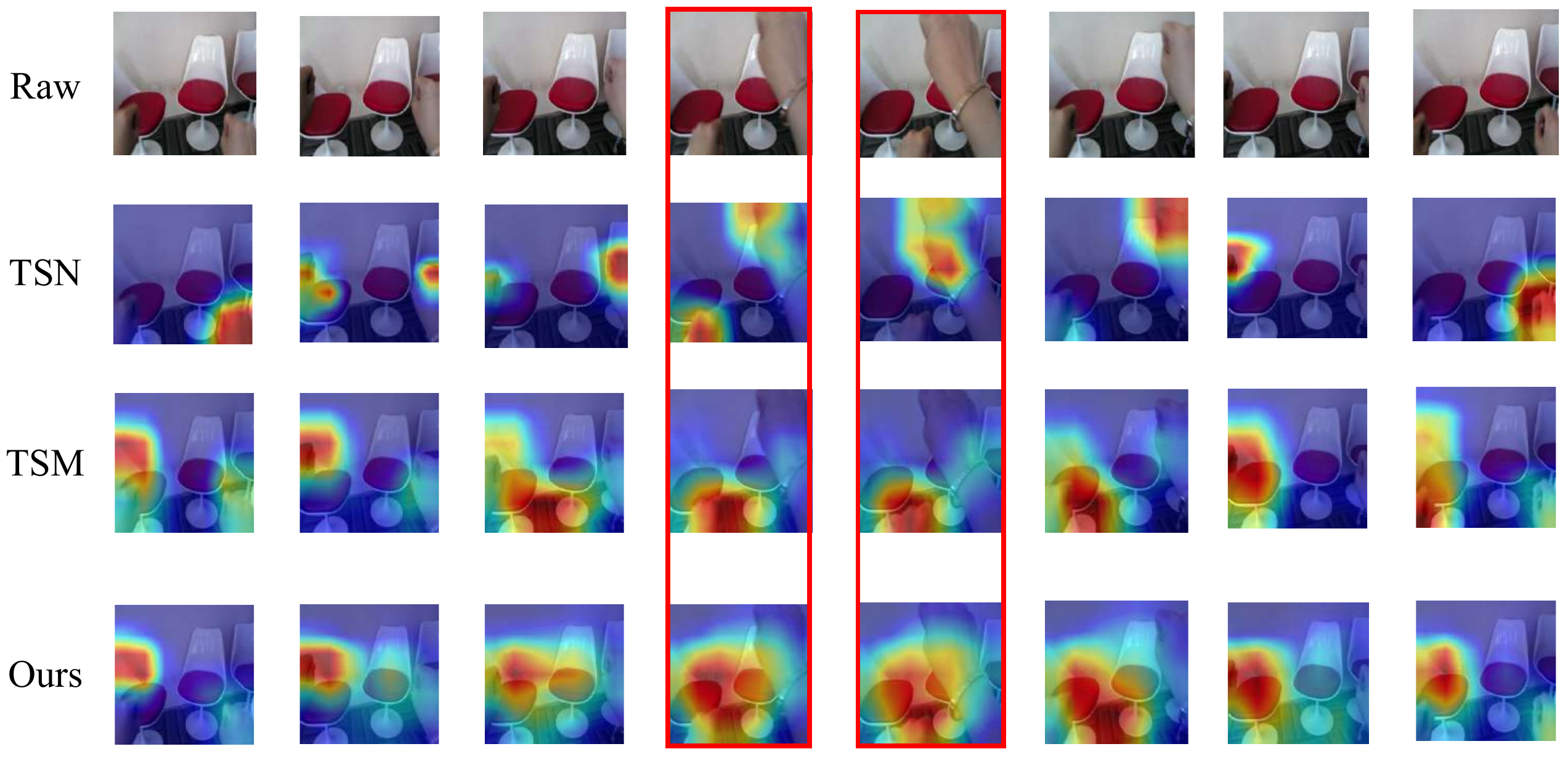}
    \caption{Visualization for significant features extracted by TSN, TSM and our ACTION-Net for the action \textit{`Rotate fists counterclockwise'}. Features extracted by each method are visualized by using CAM~\cite{zhou2016learning}. Compared to TSN and TSM, it can be noticed that ACTION-Net is able to extract features that are related to movements in an action especially for highlighted frames i.e., 4th and 5th columns. More examples can be referred to \textit{Supplementary Materials}.}
    \label{fig:heatmap}
\end{figure*}
Video understanding has attracted an increasing amount of interest, since it is a crucial step towards real-world applications, such as Virtual Reality/Augmented Reality (VR/AR) and video-sharing social networking services. For instance, millions of videos are uploaded to TikTok, Douyin, and Xigua Video to be processed everyday, wherein video understanding acts a pivotal part. However, the explosive growth in this video streaming gives rise to challenges on performing video understanding at high accuracy and low computation cost. 

Action recognition, a fundamental problem in video understanding, has been a growing demand in video-related applications, such as content moderation (i.e., recognize content in videos that break terms of service) and content recommendations (i.e., videos are ranked by most liked and recommended to similar customers). The complex actions in videos are normally temporal-dependent, which do not only contain spatial information within each frame but also include temporal information over a duration. For example, symmetric action pairs \textit{(`opening a box', `closing a box')}, \textit{(`rotate fists clockwise', `rotate fists counterclockwise')} contain similar features in spatial domains, but the temporal information is completely reversed. Traditional human action recognition is more \textit{scene-related}~\cite{soomro2012ucf101,kuehne2011hmdb,kay2017kinetics}, wherein actions are not as temporal-dependent e.g., \textit{`apply eye makeup', `walking', `running'}. With how rapid technology is developing, like VR, which requires employing gestures to interact with environments, \textit{temporal-related} action recognition has recently become a focus for research.

The mainstreams of existing methods are 3D CNN-based frameworks and 2D CNN-based frameworks. 3D CNNs have been shown to be effective in terms of spatio-temporal modeling~\cite{tran2015learning,chen20182,stroud2020d3d}, but spatio-temporal modeling is unable to capture adequate information contained in videos. The two-stream architecture was proposed to take spatio-temporal information and optical flow into account~\cite{simonyan2014two,carreira2017quo,shi2019two}, which boosted performance significantly compared to the one-stream architecture. However, computation on optical flow is very expensive, which poses challenges on real-world applications. 3D CNNs suffer from problems including overfitting and slow convergence~\cite{hara2018can}. With more large-scale datasets being released, such as Kinetics~\cite{carreira2017quo}, Moments in Time~\cite{monfort2019moments} and ActivityNet~\cite{caba2015activitynet}, optimizing 3D CNNs becomes much easier and more popular. However, heavy computations inherent in 3D CNN-based frameworks contribute to slow inferences, which would limit their deployment on real-world applications, such as VR that relies on online video recognition. Current 2D CNN-based frameworks ~\cite{karpathy2014large,simonyan2014two,wang2016temporal,zhou2018temporal,lin2019tsm} enjoy lightweight and fast inferences. These approaches operated on a sequence of short snippets (known as segments) sparsely sampled from the entire video and were initially introduced in TSN~\cite{wang2016temporal}. Original 2D CNNs lack the ability of temporal modeling, which causes losing essential sequential information in some actions e.g., \textit{`opening a box'} vs \textit{`closing a box'}. TSM~\cite{lin2019tsm} introduced temporal information to 2D CNN-based frameworks by shifting a part of channels on the temporal axis, which significantly improved the baseline for 2D CNN-based frameworks. However, TSM still lacks explicit temporal modeling for an action, such as motion information. Recent works~\cite{lee2018motion,jiang2019stm,li2020tea,liu2020teinet,liu2020tam} introduced embedded modules into 2D CNNs in terms of ResNet architecture~\cite{he2016deep}, which possessed the capability for motion modeling. In order to capture multi-type information contained by videos, previous works normally operated on input-level frames. For instance, SlowFast networks sampled raw videos at multiple rates to characterize slow and fast actions; two-stream networks utilized pre-computed optical flow for reasoning motion information. This kind of approaches commonly require multi-branch networks, which need expensive computations. 

Inspired by the aforementioned observation, we propose a new \textit{plug-and-play} and \textit{lightweight} sp\textbf{A}tio-temporal, \textbf{C}hannel and mo\textbf{T}ion excitat\textbf{ION} (ACTION) module to effectively process the multi-type information on the feature level inside a single network by adopting multipath excitation. The combination of spatio-temporal features and motion features can be understood similarly as the two-stream architecture~\cite{simonyan2014two}, but we model the motion inside the network based on the feature level rather than generating another type of input (e.g., optical flow~\cite{ilg2017flownet}) for training the network, which significantly reduces computations. Inspired by SENet, the channel-wise features are extracted based on the temporal domain to characterize the channel interdependencies for the network. Correspondingly, a neural architecture equipped with such a module is dubbed ACTION-Net. ACTION comprises three components for extracting aforementioned features (1) Spatio-Temporal Excitation (STE), (2) Channel Excitation (CE) and (3) Motion Excitation (ME). Figure~\ref{fig:heatmap} visualizes features characterized by TSN, TSM, and ACTION-Net for the action \textit{`rotate fists counterclockwise'}. It can be observed that both TSN and TSM focus on recognizing objects (two fists) independently instead of reasoning an action. Compared to TSN and TSM, our proposed ACTION-Net better characterizes an action by representing feature maps that cover the two fists, especially for the highlighted 4th and 5th columns. In a nutshell, our contributions are three-fold:
\begin{itemize}
    \item We propose an ACTION module that works in a \textit{plug-and-play} manner, which is able to extract appropriate spatio-temporal patterns, channel-wise features, and motion information to recognize actions.
    \item A simple yet effective neural architecture referred to ACTION-Net, which we demonstrate on three backbones i.e., ResNet-50~\cite{he2016deep}, BNIception~\cite{ioffe2015batch} and MobileNet V2~\cite{sandler2018mobilenetv2}.
    \item We have conducted extensive experiments and shown our superior performances on three datasets Something-Something V2~\cite{goyal2017something}, Jester~\cite{materzynska2019jester} and EgoGesture~\cite{zhang2018egogesture}. 
\end{itemize}

\section{Related Works}
In this section, we discuss related works by taking 2D and 3D CNN-based frameworks into account, wherein SENet~\cite{hu2018squeeze} and TEA~\cite{li2020tea} inspired us to propose the ACTION-Net.  

\subsection{3D CNN-based Framework}
The 3D CNN-based framework has a spatio-temporal modeling capability, which enhances model performance for video action recognition~\cite{tran2015learning,hara2018can,tran2015learning}. I3D~\cite{carreira2017quo} inflated the ImageNet pre-trained 2D kernels to 3D kernels for capturing spatio-temporal information. To better represent motion patterns, I3D utilized pre-computed optical flow together with RGB (also known as the two-stream architecture). SlowFast networks~\cite{feichtenhofer2019slowfast} were proposed to handle inconsistent speeds of actions in videos e.g., running and walking, which involved a slow branch and a fast branch to model slow actions and fast actions, respectively. Although 3D CNN-based approaches have achieved exciting results on several benchmark datasets, they contain massive parameters. In this case, various problems are caused, such as easily overfitting~\cite{hara2018can} and difficulty in converging~\cite{tran2018closer}, which pose challenges including ineffective inferences for online streaming videos in real-world applications. Although recent works~\cite{qiu2017learning,tran2018closer} have demonstrated that 3D convolution can be factorized to lessen computations to some extent, the computation is still much more of a burden when compared to 2D CNN-based frameworks.

\subsection{2D CNN-based Framework}
TSN~\cite{wang2016temporal} was the first proposed framework that applied 2D CNNs for video action recognition, which introduced the concept of \textit{`segment'} to process videos i.e., extract short snippets over a long video sequence with a uniform sparse sampling scheme. However, the direct use of 2D CNNs lacks temporal modeling for video sequences. TSM~\cite{lin2019tsm} firstly introduced temporal modeling to 2D CNN-based frameworks, in which a shift operation for a part of channels was embedded into 2D CNNs. However, TSM lacks explicit temporal modeling for actions such as differences among neighbouring frames. Recently, several works have proposed modules to be embedded into 2D CNNs. These modules are able to model the motion and temporal information. For instance, MFNet~\cite{lee2018motion}, TEINet~\cite{liu2020teinet} and TEA~\cite{li2020tea}, which introduced this type of  modules, were demonstrated to be effective on the ResNet architecture. STM~\cite{jiang2019stm} proposed a block for modeling the spatio-temporal and motion information instead of the ordinary residual block. GSM~\cite{sudhakaran2020gate} leverages group spatial gating to control interactions in spatial-temporal decomposition.

\subsection{SENet and Beyond}
Hu~\etal~\cite{hu2018squeeze} introduced a SENet architecture. A squeeze-and-excitation (SE) block was proposed to be embedded into 2D CNNs. In this case, the learning of channel-wise features regarding image recognition tasks was enhanced by explicitly modeling channel interdependencies. To tackle this, the SE block utilized two fully connected (FC) layers in a squeeze-and-unsqueeze manner then applied a Sigmoid activation for exciting essential channel-wise features. However, it processes each image independently without considering critical information such as temporal properties for videos. To tackle this issue, TEA~\cite{li2020tea} introduces motion excitation (ME) and multiple temporal aggregation (MTA) in tandem to capture short- and long-range temporal evolution. It should be noted that MTA is specifically designed for Res2Net~\cite{gao2019res2net}, which means TEA can only be embedded into Res2Net. Inspired by these two previous works, we first propose STE and CE modules beyond the SE module by addressing the spatio-temporal perspective and channel interdependencies in temporal dimension. The ACTION module is then constructed by assembling STE, CE and ME in a parallel manner, in which case, multi-type information in videos can be activated.

\section{Design of ACTION}
In this section, we are going to introduce technical details for our proposed ACTION-Net together with ACTION module. As the ACTION module consists of three sub-modules i.e., Spatio-Temporal Excitation (STE), Channel Excitation (CE) and Motion Excitation (ME), we firstly introduce these three sub-modules respectively and then give an overview on how to integrate them to an ACTION module. Notations used in this section are: $N$ (batch size), $T$ (number of segments), $C$ (channels), $H$ (height), $W$ (width) and $r$ (channel reduce ratio). It should be noticed that all tensors outside the ACTION module are 4D i.e., (N$\times$T, C, H, W). We first reshape the input 4D tensor to 5D tensor (N, T, C, H, W) before feeding to the ACTION in order to enable the operation on specific dimension inside the ACTION. The 5D output tensor is then reshaped to 4D before being fed to the next 2D convolutional block. 
\begin{figure}[!htbp]
    \centering
    \subfigure[]{
        \includegraphics[width=.18\textwidth]{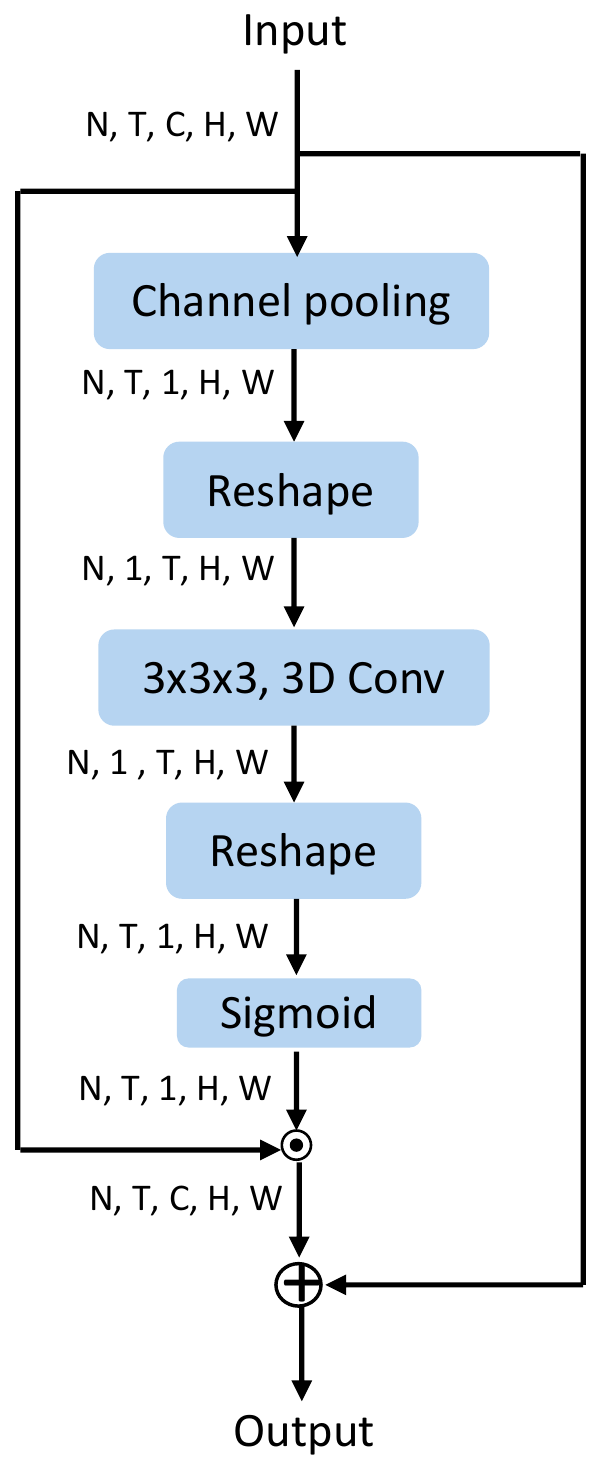}
        \label{fig:three_paths_a}
    }
    \hspace{20pt}
    \subfigure[]{
        \includegraphics[width=.15\textwidth]{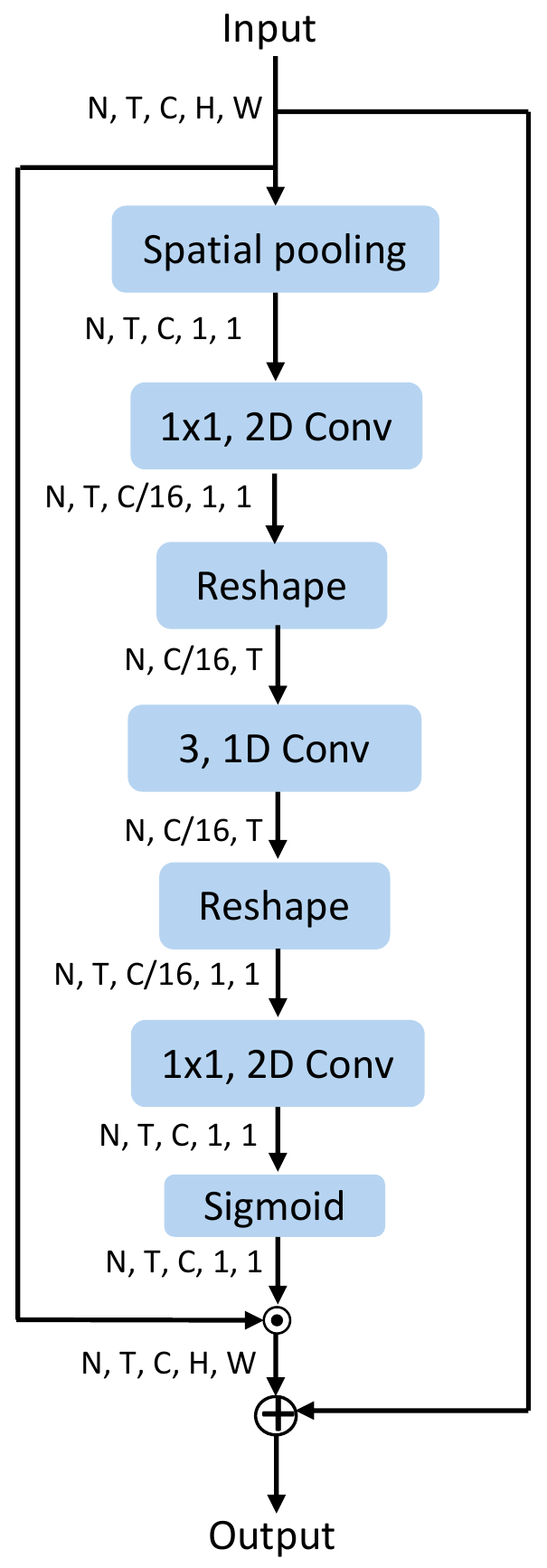}
        \label{fig:three_paths_b}
    }    
    \subfigure[]{
        \includegraphics[width=.35\textwidth]{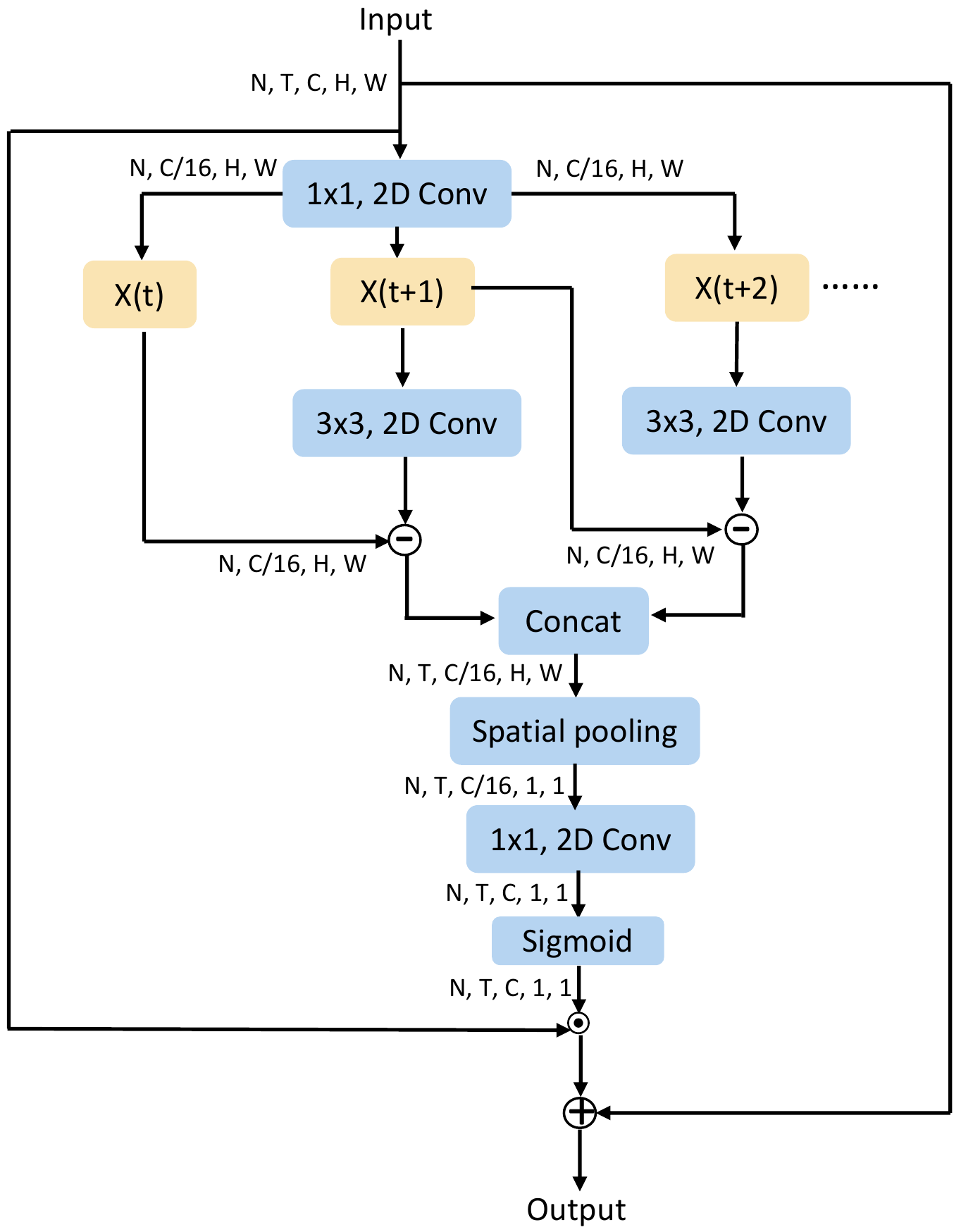}
        \label{fig:three_paths_c}
    }    
    \caption{ACTION module consists of three sub-modules (a) Spatio-Temporal Excitation (STE) module, (b) Channel Excitation (CE) module and (c) Motion Excitation (ME) module.}
    \label{fig:three_paths}
\end{figure}

\begin{figure*}[!htbp]
    \centering
    \includegraphics[width=.9\textwidth]{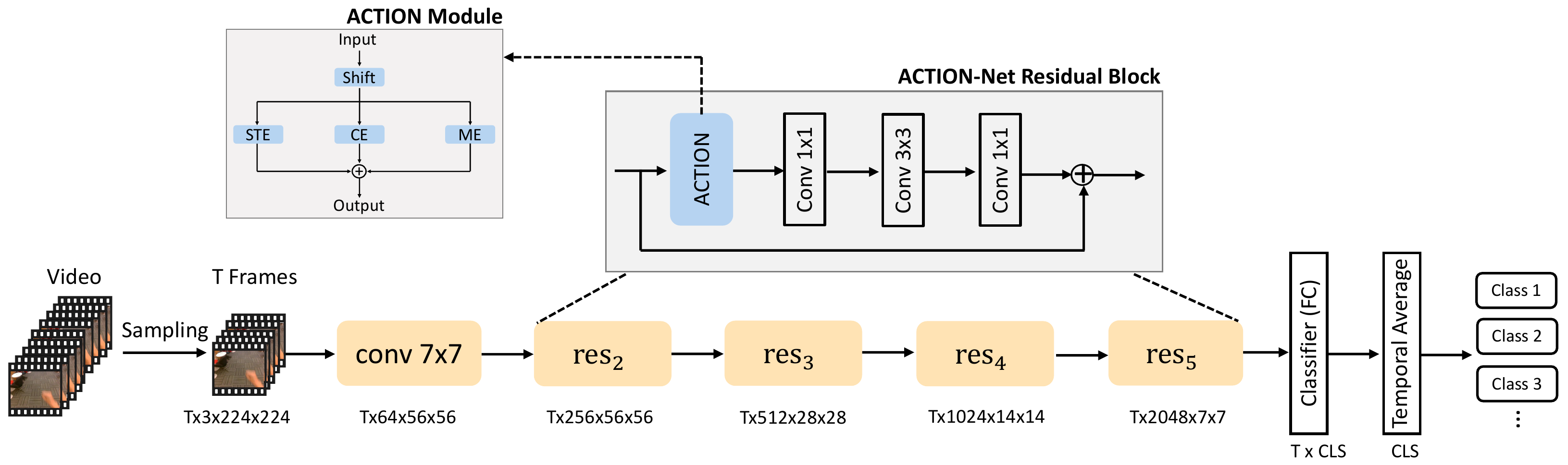}
    \caption{ACTION-Net architecture for ResNet-50~\cite{he2016deep}. The size of output feature map is given for each layer ($CLS$ refers to number of classes and $T$ refers to number of segments). The input video is firstly split into T segments equally and then one frame from each segment is randomly sampled~\cite{wang2016temporal}. The ACTION module is inserted at the start in each residual block. Performance of different embedded locations can be referred to \textit{Supplementary Materials}~\ref{sup-sec:location}.}
    \label{fig:backbone}
\end{figure*}

\subsection{Spatio-Temporal Excitation (STE)}
STE is efficiently designed for exciting spatio-temporal information by utilizing 3D convolution. To achieve this,  STE generates a spatio-temporal mask $\mathbf{M}\in\mathbb{R}^{N\times T\times 1 \times H \times W}$ that is used for element-wise multiplying the input $\mathbf{X}\in\mathbb{R}^{N\times T\times C \times H \times W}$ across all channels. As illustrated in Fig.~\ref{fig:three_paths_a}, given an input $\mathbf{X}\in\mathbb{R}^{N\times T\times C \times H \times W}$, we first average the input tensor across channels in order to get a global spatio-temporal tensor $\mathbf{F}\in \mathbb{R}^{N\times T\times 1 \times H \times W}$ with respect to the channel axis. Then we reshape $\mathbf{F}$ to $\mathbf{F}^*\in \mathbb{R}^{N\times 1\times T \times H \times W}$ to be fed to a 3D convolutional layer $\mathbf{K}$ with kernel size $3\times3\times3$, which can be formulated as 
\begin{equation}
    \mathbf{F}_o^* = \mathbf{K} * \mathbf{F}^*
\end{equation}
We finally reshape $\mathbf{F}_o^*$ back to $\mathbf{F}_o\in \mathbb{R}^{N\times T\times 1 \times H \times W}$ and feed it to a Sigmoid activation in order to get the mask $\mathbf{M}\in \mathbb{R}^{N\times T\times 1 \times H \times W}$, which can be represented as
\begin{equation}
    \mathbf{M} = \delta(\mathbf{F}_o)
\end{equation}
The final output can be interpreted as
\begin{equation}\label{eq:sub_module_output}
    \mathbf{Y} = \mathbf{X} + \mathbf{X} \odot \mathbf{M} 
\end{equation}
Compared to the conventional 3D convolutional operation, STE is much more computationally efficient as the input feature $\mathbf{F}^*$ is averaged across channels. Each channel of the input tensor $\mathbf{X}$ can perceive the importance of spatio-temporal information from a refined feature excitation $\delta(\mathbf{F}_o)$.

\subsection{Channel Excitation (CE)}
CE is designed similarly to SE block~\cite{hu2018squeeze} as shown in Fig.~\ref{fig:three_paths_b}. The main difference between CE and SE is that we insert a 1D convolutional layer between two FC layers to characterize temporal information for channel-wise features. Concretely, given an input $\mathbf{X}\in\mathbb{R}^{N\times T\times C \times H \times W}$, we firstly access the global spatial information of the input feature by spatial average pooling the input, which can be represented as
\begin{equation}
    \mathbf{F} = \frac{1}{H\times W}\sum^{H}_{i=1}\sum^{W}_{j=1}\mathbf{X[:,:,:,i,j]}
    \label{eq:spatial_pool}
\end{equation}
where $\mathbf{F}\in \mathbb{R}^{N\times T\times C \times 1 \times 1}$. We squeeze the number of channels for $\mathbf{F}$ by a scale ratio $r$ ($r = 16$ in this work), which can be interpreted as 
\begin{equation}
    \mathbf{F}_r = \mathbf{K}_1 * \mathbf{F}
    \label{eq:squeeze}
\end{equation}
where $\mathbf{K}_1$ is a $1\times1$ 2D convolutional layer and $\mathbf{F}_r\in \mathbb{R}^{N\times T\times \frac{C}{r} \times 1 \times 1}$. We then reshape $\mathbf{F}_r$ to $\mathbf{F}_r^*\in \mathbb{R}^{N\times \frac{C}{r} \times T \times 1 \times 1}$ to enable the temporal reasoning. A 1D convolutional layer $\mathbf{K}_2$ with kernel size 3 is utilized to process $\mathbf{F}_r^*$ as
\begin{equation}
    \mathbf{F}^*_{temp} = \mathbf{K}_2 * \mathbf{F}_r^*
\end{equation}
where $\mathbf{F}^*_{temp}\in \mathbb{R}^{N\times \frac{C}{r} \times T \times 1 \times 1}$. $\mathbf{F}^*_{temp}$ is then reshaped to $\mathbf{F}_{temp}\in \mathbb{R}^{N\times T \times \frac{C}{r} \times 1 \times 1}$, which is then unsqueezed by using a $1\times 1$ 2D convolutional layer $\mathbf{K}_3$ and fed to a Sigmoid activation. These are the last two steps to obtain the channel mask $\mathbf{M}$, which can be formulated respectively
\begin{equation}
    \mathbf{F}_o = \mathbf{K}_3 * \mathbf{F}_{temp}
    \label{eq:unsqueeze}
\end{equation}
\begin{equation}
    \mathbf{M} = \delta(\mathbf{F}_o)
    \label{eq:mask}
\end{equation}
where $\mathbf{F}_o\in \mathbb{R}^{N\times T \times {C} \times 1 \times 1}$ and $\mathbf{M}\in \mathbb{R}^{N\times T \times {C} \times 1 \times 1}$. Finally, the output of CE is formulated as the same as in equation~\eqref{eq:sub_module_output} using the new generated mask.

\subsection{Motion Excitation (ME)}
ME has been explored by~\cite{jiang2019stm,li2020tea} previously, which aims to model motion information based on the feature level instead of the pixel level. Different from previous work~\cite{jiang2019stm,li2020tea} that proposed a whole block for extracting motion, we use the ME in parallel with two modules mentioned in previous two sections. As illustrated in Fig.~\ref{fig:three_paths_c}, the motion information is modeled by adjacent frames. We adopt the same squeeze and unsqueeze strategy as used in the CE sub-module by using two $1\times 1$ 2D convolutional layers, which can be referred to equation~\eqref{eq:squeeze} and equation~\eqref{eq:unsqueeze} respectively. Given the feature $\mathbf{F}_r\in \mathbb{R}^{N\times T \times \frac{C}{r} \times H \times W}$ processed by the squeeze operation, motion feature is modeled following the similar operation presented in~\cite{jiang2019stm,li2020tea}, which can be represented as 
\begin{equation}
    \mathbf{F}_m = \mathbf{K} * \mathbf{F}_r[:,t+1,:,:,:] - \mathbf{F}_r[:,t,:,:,:]
\end{equation}
where $\mathbf{K}$ is a $3\times 3$ 2D convolutional layer and $\mathbf{F}_m\in \mathbb{R}^{N\times 1 \times \frac{C}{r} \times H \times W}$. The motion feature is then concatenated to each other according to the temporal dimension and 0 is padded to the last element i.e., $\mathbf{F}_M = [\mathbf{F}_m(1), \dots, \mathbf{F}_m(t-1), 0], \mathbf{F}_M\in \mathbb{R}^{N\times T \times \frac{C}{r} \times H \times W}$. The $\mathbf{F}_M$ is then processed by spatial average pooling same as in equation~\eqref{eq:spatial_pool}. The feature output $\mathbf{F}_o\in \mathbb{R}^{N\times T \times C\times 1 \times 1}$ and the mask $\mathbf{M}\in \mathbb{R}^{N\times T \times {C} \times 1 \times 1}$ can then be achieved similarly as in CE through equation~\eqref{eq:unsqueeze} and equation~\eqref{eq:mask} respectively.

\subsection{ACTION-Net}
The overall ACTION module takes the element-wise addition of three excited features generated by STE, CE and ME respectively (see ACTION module block in Fig.~\ref{fig:backbone}). By doing this, the output of the ACTION module can perceive information from a spatio-temporal perspective, channel interdependencies and motion. Figure~\ref{fig:backbone} shows the ACTION-Net architecture for ResNet-50, wherein the ACTION module is inserted at the beginning in each residual block. It does not require any modification for original components in the block. Details of ACTION-Net architectures for MobileNet V2 and BNInception can be referred to \textit{Supplementary Materials}.

\section{Experiments}
We first show that ACTION-Net is able to consistently improve the performance for 2D CNNs compared to previous two fundamental works TSN~\cite{wang2016temporal} and TSM~\cite{lin2019tsm} on three datasets i.e., EgoGesture~\cite{zhang2018egogesture}, Something-Something V2~\cite{goyal2017something} and Jester~\cite{materzynska2019jester}. We then perform extensive experiments for comparing ACTION-Net with state-of-the-arts on these three datasets. Abundant ablation studies are conducted to analyze the efficacy for each excitation path in ACTION-Net. Finally, we further compare ACTION-Net with TSM on three backbones i.e., ResNet-50, MobileNet V2 and BNInception by considering performance and efficiency ($\eta$).

\subsection{Datasets}
We evaluated the performance for the proposed ACTION-Net on three large-scale and widely used action recognition datasets i.e., Something-Something V2~\cite{goyal2017something}, Jester~\cite{materzynska2019jester} and EgoGesture~\cite{zhang2018egogesture}. The Something-Something V2 dataset~\cite{lee2018motion,liu2020teinet,yang2020temporal,zolfaghari2018eco} is a large collection of humans performing actions with everyday objects. It includes 174 categories with 168,913 training videos, 24,777 validation videos and 27,157 testing videos. Jester~\cite{jiang2019stm,kopuklu2019real,zhang2020temporal,materzynska2020something,kopuklu2020drivermhg} is a third-person view gesture dataset, which has a potential usage for human computer interaction. It has 27 categories with 118,562 training videos, 14,787 validation videos and 14,743 testing videos. EgoGesture~\cite{kopuklu2019real,wang2020catnet,shi2019two,shi2019skeleton,abavisani2019improving,shi2020skeleton} is a large-scale dataset for egocentric hand gesture recognition recorded by a head-mounted camera, which is designed for VR/AR use cases. It involves 83 classes of gestures with 14,416 training samples, 4,768 validation samples and 4,977 testing samples.

\subsection{Implementation Details}
\noindent \textbf{Training.}\hspace{2pt} We conducted our experiments on video action recognition tasks by following the same strategy mentioned in TSN~\cite{wang2016temporal}. Given an input video, we firstly divided it into $T$ segments of equal duration. Then we randomly selected one frame from each segment to obtain a clip with $T$ frames. The size of the shorter side of these frames is fixed to 256 and corner cropping and scale-jittering were ultilized for data augmentation. Each cropped frame was finally resized to $224\times 224$, which was used for training the model. The input fed to the model is of the size $N\times T \times 3 \times 224 \times 224$, in which $N$ is the batch size, $T$ is the number of segments. 

The models were trained on a NVIDIA DGX station with four Tesla V100 GPUs. We adopted SGD as optimizer with a momentum of 0.9 and a weight decay of $5\times 10^{-4}$. Batch size was set as $N=64$ when $T=8$ and $N=48$ when $T=16$. Network weights were initialized using ImageNet pretrained weights. For Something-Something V2, we started with a learning rate of 0.01 and reduced it by a factor of 10 at 30, 40, 45 epochs and stopped at 50 epochs. For Jester dataset, we started with a learning rate of 0.01 and reduced it by a factor of 10 at 10, 20, 25 epochs and stopped at 30 epochs. For EgoGesture dataset, we started with a learning rate of 0.01 and reduced it by a factor of 10 at 5, 10, 15 epochs and stopped at 25 epochs.

\noindent \textbf{Inference.}\hspace{2pt} We utilized the three-crop strategy following \cite{jiang2019stm,wang2018non,feichtenhofer2019slowfast} for inference. We firstly scaled the shorter side to 256 for each frame and took three crops of $256\times 256$ from scaled frames. We randomly sampled from the full-length video for 10 times. The final prediction was the averaged Softmax score for all clips. 

\subsection{Improving Performance of 2D CNNs}
We compare ACTION-Net to two fundamental 2D CNN counterparts TSN and TSM. As illustrated in Table~\ref{tab:2D_baseline_compare}, 
\begin{table}[!htbp]
    \centering
    \caption{ACTION-Net consistently outperforms 2D counterparts on three representative datasets. All methods use ResNet-50 as backbone and 8 frames input for the fair comparison.}
    \setlength{\tabcolsep}{0.45mm}{
    \begin{threeparttable}
    \begin{tabular}{cccccc}
        \toprule
         Dataset& Model& Top-1 & $\triangle$Top-1& Top-5& $\triangle$Top-5\\ 
         \midrule
         \multirow{3}{*}{EgoGesture\tnote{*}}& TSN & 83.1& -& 97.3&- \\
                                    & TSM & 92.1& + 9.0&98.3  &+ 1.0\\ 
                                    & \cellcolor{columbiablue}ACTION-Net & \cellcolor{columbiablue}94.2&\cellcolor{columbiablue} + 11.1&\cellcolor{columbiablue}98.7 &\cellcolor{columbiablue}+ 1.4 \\\hline
         \multirow{3}{*}{SomethingV2}& TSN & 27.8& -& 57.6&-\\
                                    & TSM &58.7& + 30.9& 84.8 &+ 27.2\\ 
                                    & \cellcolor{columbiablue}ACTION-Net &\cellcolor{columbiablue}62.5&\cellcolor{columbiablue} + 34.7& \cellcolor{columbiablue}87.3&\cellcolor{columbiablue}+ 29.7\\ \hline  
         \multirow{3}{*}{Jester}& TSN & 81.0& -& 99.0&-\\
                                & TSM& 94.4& + 13.4& 99.7&+ 0.7\\
                                & \cellcolor{columbiablue}ACTION-Net& \cellcolor{columbiablue}97.1& \cellcolor{columbiablue}+ 16.1& \cellcolor{columbiablue}99.8&\cellcolor{columbiablue}+ 0.8\\
         \bottomrule
    \end{tabular}
   \begin{tablenotes}
     \item[*] We re-implement TSN and TSM using the official public code in~\cite{lin2019tsm}.
   \end{tablenotes}
    \end{threeparttable}
    \label{tab:2D_baseline_compare}
    }
\end{table}
ACTION-Net consistently outperforms these two 2D CNN counterparts on all three datasets. It is worth nothing that TSN does not contain any component that is able to model the temporal information. By employing a temporal shift operation to a part of channels, TSM introduces some temporal information to the network, which significantly improves the 2D CNN baseline compared to TSN. However, TSM still lacks explicit temporal modeling. By adding the ACTION module to TSN, ACTION-Net takes spatio-temporal modeling, channel interdependencies modeling and motion modeling into account. It can be noticed that the Top-1 accuracy of ACTION-Net is improved by 2\%, 3.8\% and 2.7\% compared to TSM with respect to EgoGesture, Something-Something V2 and Jester datasets.

\subsection{Comparisons with the State-of-the-Art}
\begin{table*}[!htbp]
    \centering
    \caption{Comparisons with the state-of-the-arts on Something-Something V2, Jester and EgoGesture datasets.}
    \label{tab:SOTAs_compare}
    \setlength{\tabcolsep}{.9mm}{
    \begin{threeparttable}
    \begin{tabular}{c|c|ccccccccc}
        \toprule
         \multicolumn{1}{c}{\multirow{2}{*}{Method}}&\multicolumn{1}{c}{\multirow{2}{*}{Backbone}}& \multirow{2}{*}{Plug-and-play}& \multirow{2}{*}{Pretrain}& \multirow{2}{*}{Frame}&
         \multicolumn{2}{c}{Something V2}&
         \multicolumn{2}{c}{Jester}&
         \multicolumn{2}{c}{EgoGesture}
         \\ \cline{6-7} \cline{8-9} \cline{9-11}
         \multicolumn{1}{c}{}&\multicolumn{1}{c}{}&\multicolumn{1}{c}{}&&&Top-1 & Top-5& Top-1 & Top-5& Top-1 & Top-5\\ 
         \midrule
         C3D + RSTTM~\cite{zhang2018egogesture}&  {-}& & {Scratch}
         & 16& -& - & -& -& 89.3& -\\ \cline{0-0}\cline{2-11}
         C3D~\cite{kopuklu2019real}& {ResNext-101}& & {Kinetics}
         & 16& -& - & 95.9& -& 90.9& -\\ \hline
         TRN Multiscale~\cite{lin2019tsm}& BNInception& & ImageNet
         & 8& 48.8& 77.6 & 95.3& - & - & -\\ \cline{0-0}\cline{2-11}
         MFNet-C50~\cite{lee2018motion}&\multirow{16}{*}{ResNet-50}& & {ImageNet}
         & 7& -& - & 96.1& 99.7& - & -\\ \cline{0-0}\cline{3-11}
         \multirow{2}{*}{TSN~\cite{lin2019tsm}}& \multirow{2}{*}{}& \multirow{2}{*}\checkmark& \multirow{2}{*}{Kinetics}
         & 8& 27.8& 57.6 & 81.0 & 99.0 & 83.1 & 98.3\\
         &&&&16& 30.0&60.5 & 82.3& 99.2& -&-\\ \cline{0-0}\cline{3-11}
         \multirow{2}{*}{TSM~\cite{jiang2019stm}}& \multirow{2}{*}{}& \multirow{2}{*}{\checkmark}& \multirow{2}{*}{Kinetics}
         & 8& 58.7& 84.8& 94.4& 99.7& 92.1& 98.3\\
         &&&& 16& 61.2&  86.9& 95.3& 99.8& -&-\\
        \cline{0-0}\cline{3-11}
        \multirow{2}{*}{GST~\cite{luo2019grouped}}& \multirow{2}{*}{}& & \multirow{2}{*}{ImageNet}
                         & 8& 61.6& 87.2 & - & - & - & -\\
                         &&&&16& 62.6&87.9 & -& -& -&-\\ \cline{0-0}\cline{3-11}
        bLVNet-TAM~\cite{liu2020teinet}& & & ImageNet
                                 & 32& 61.7& 88.1 & - & - & - & -\\\cline{0-0}\cline{3-11}
        CPNet~\cite{liu2019learning}& & & ImageNet
                                 & 24& 57.7& 84.0 & - & - & - & -\\\cline{0-0}\cline{3-11}
        \multirow{2}{*}{TEINet~\cite{liu2020teinet}}&  \multirow{2}{*}{}& \multirow{2}{*}\checkmark& \multirow{2}{*}{ImageNet}
                                     & 8& 62.7& - & - & - & - & -\\
                                     &&&&16& 63.0&- & -& -& -&-\\ \cline{0-0}\cline{3-11}
        \multirow{2}{*}{STM~\cite{jiang2019stm}}& \multirow{2}{*}{}& & \multirow{2}{*}{ImageNet}
                 & 8& 62.3& 88.8 & 96.6 & 99.9 & - & -\\
                 &&&&16& 64.2&89.8 & 96.7& 99.9& -&-\\ \cline{0-0}\cline{3-11}
        \multirow{2}{*}{TEA~\cite{li2020tea}}& \multirow{2}{*}{}& & \multirow{2}{*}{ImageNet}
                 & 8& -& - & 96.5& 99.8& 92.3 & 98.3\\
                 &&&&16& \textbf{64.5}&\textbf{89.8} & 96.7& 99.8& 92.5&98.5\\ 
         \rowcolor{columbiablue}&& & 
         & 8& 62.5& 87.3& 97.1& 99.8& 94.2&98.7\\
         \rowcolor{columbiablue}\multirow{-2}{*}{ACTION-Net\tnote{1}}&&\multirow{-2}{*}\checkmark&\multirow{-2}{*}{ImageNet}&16& 64.0&89.3& \textbf{97.1}& \textbf{99.9}& \textbf{94.4}&\textbf{98.8}\\         
         \bottomrule
    \end{tabular}
   \begin{tablenotes}
     \item[1] ACTION is inserted into each residual block in this experiment. 
   \end{tablenotes}
    \end{threeparttable}
    }
\end{table*}
We compare our approach with the state-of-the-art on Something-Something V2, Jester and EgoGesture, which is summarized in Table~\ref{tab:SOTAs_compare}. We mainly compare our approach with 2D CNN counterparts as 3D CNN-based frameworks are more favored in \textit{scene-related} datasets e.g., Kinetics~\cite{qiu2017learning,feichtenhofer2019slowfast}. The superiority of ACTION-Net on Jester and EgoGesture is quite impressive. It is clear that even ACTION-Net with 8 RGB frames as input achieves the state-of-the-art performance compared to other methods, which confirms the remarkable ability of ACTION-Net for integrating useful information from three excitation paths. In terms of Something-Something V2, ACTION-Net also achieves competitive results compared to STM and TEA. It should be noted that both STM and TEA are specifically designed for ResNet and Res2Net respectively, while our ACTION enjoys being easily equipped by other architectures i.e., MobileNet V2 and BNInception investigated in this work.

\subsection{Ablation Study}
\begin{table*}[!htbp]
    \centering
    \hspace{-0.5cm}
    \begin{minipage}{0.6\textwidth}
        \caption{Accuracy and model complexity on EgoGesture dataset. Three excitation types are compared to TSM and TSN. All methods use ResNet-50 as backbone and 8 frames input for fair comparison. The least FLOPs/$\triangle$FLOPs/Param. and the best performance for ACTION-Net and sub-modules are highlighted as bold.}
        \begin{tabular}{cccccc}
            \toprule
             \multicolumn{2}{c}{Method}& FLOPs& $\triangle$FLOPs& Param.& Top-1 \\ 
             \midrule
             \multicolumn{2}{c}{TSN} &  33G& -& 23.68M& 83.1\\ 
             \multicolumn{2}{c}{TSM} &  33G& -& 23.68M& 92.1\\ \hline \hline
             \multirow{4}{*}{Ours} &STE &  \textbf{33.1G}& \textbf{+0.1G (+0.3\%)}& \textbf{23.9M}& 93.8\\ 
             &CE &  33.16G& +0.16G (+0.5\%)& 26.08M& 93.8\\ 
             &ME &  34.69G& +1.69G (+5.1\%)& 25.9M& 93.9\\
             &ACTION-Net &  34.75G& +1.75G (+5.3\%)& 28.08M& \textbf{94.2}\\ 
             \bottomrule
        \end{tabular}
        \label{tab:three_paths}
    \end{minipage}
    \hspace{20pt}
    \centering
    \begin{minipage}{0.3\textwidth}
        \caption{Ablation study of having ACTION included or not in each residual block stage regarding ResNet-50 backbone on EgoGesture dataset by using 8 frames input. More ACTION engaged yields better performance.}
        \begin{tabular}{cccc}
             \toprule    
             Stage &  Top-1& Top-5 \\ \hline
             \multicolumn{1}{l}{$\text{res}_2$}& 92.3& 98.2\\
             \multicolumn{1}{l}{$\text{res}_{2,3}$}& 92.9& 98.5\\
             \multicolumn{1}{l}{$\text{res}_{2,3,4}$}& 93.1& 98.5\\
             \multicolumn{1}{l}{$\text{res}_{2,3,4,5}$}& \textbf{94.2}& \textbf{98.7}\\
             \bottomrule
        \end{tabular}
        \label{tab:block_num}
    \end{minipage}
\end{table*}
\begin{table*}[!htbp]
    \centering
    \caption{ACTION-Net generalizes well across backbones and datasets (TSM is used as a baseline). Accuracy and model complexity on three backbones using Something-Something V2, Jester and EgoGesture with 8 frames as input. The most significant improvements on accuracy and the least extra FLOPs are highlighted as bold.}
    \begin{tabular}{ccccccc}
        \toprule
         Backbone& Method & FLOPs& Param.& Something V2& Jester& EgoGesture \\ 
         \midrule
         \multirow{2}{*}{ResNet-50} & TSM & 33G& 23.68M& 58.7& 94.4& 92.1\\ 
                                    & \cellcolor{columbiablue}ACTION-Net & \cellcolor{columbiablue}34.75G (+5.3\%)& \cellcolor{columbiablue}28.08M& \cellcolor{columbiablue}\textbf{62.5 (+3.8)}& \cellcolor{columbiablue}\textbf{97.1 (+2.7)}&\cellcolor{columbiablue}\textbf{94.2 (+2.1)}\\\hline
         \multirow{2}{*}{BNIception}& TSM & 16.39G& 10.36M& 53.5& 94.6&92.3\\ 
                                    & \cellcolor{columbiablue}ACTION-Net &\cellcolor{columbiablue}16.94G (+3.4\%)& \cellcolor{columbiablue}11.59M& \cellcolor{columbiablue}56.7 (+3.2)& \cellcolor{columbiablue}96.6 (+2.0)&\cellcolor{columbiablue}93.2 (+0.9)\\ \hline  
         \multirow{2}{*}{MobileNet V2}& TSM& 2.55G& 2.33M& 54.9& 95.0&92.4\\
                                & \cellcolor{columbiablue}ACTION-Net& \cellcolor{columbiablue}\textbf{2.57G (+0.8\%)}& \cellcolor{columbiablue}2.36M& \cellcolor{columbiablue}58.5 (+3.6)& \cellcolor{columbiablue}96.7 (+1.7)&\cellcolor{columbiablue}93.5 (+1.1)\\
         \bottomrule
    \end{tabular}   
    \label{tab:TSM_STCME}
\end{table*}
\begin{figure*}
    \subfigure[EgoGesture]{
        \centering
        \includegraphics[width=0.32\textwidth]{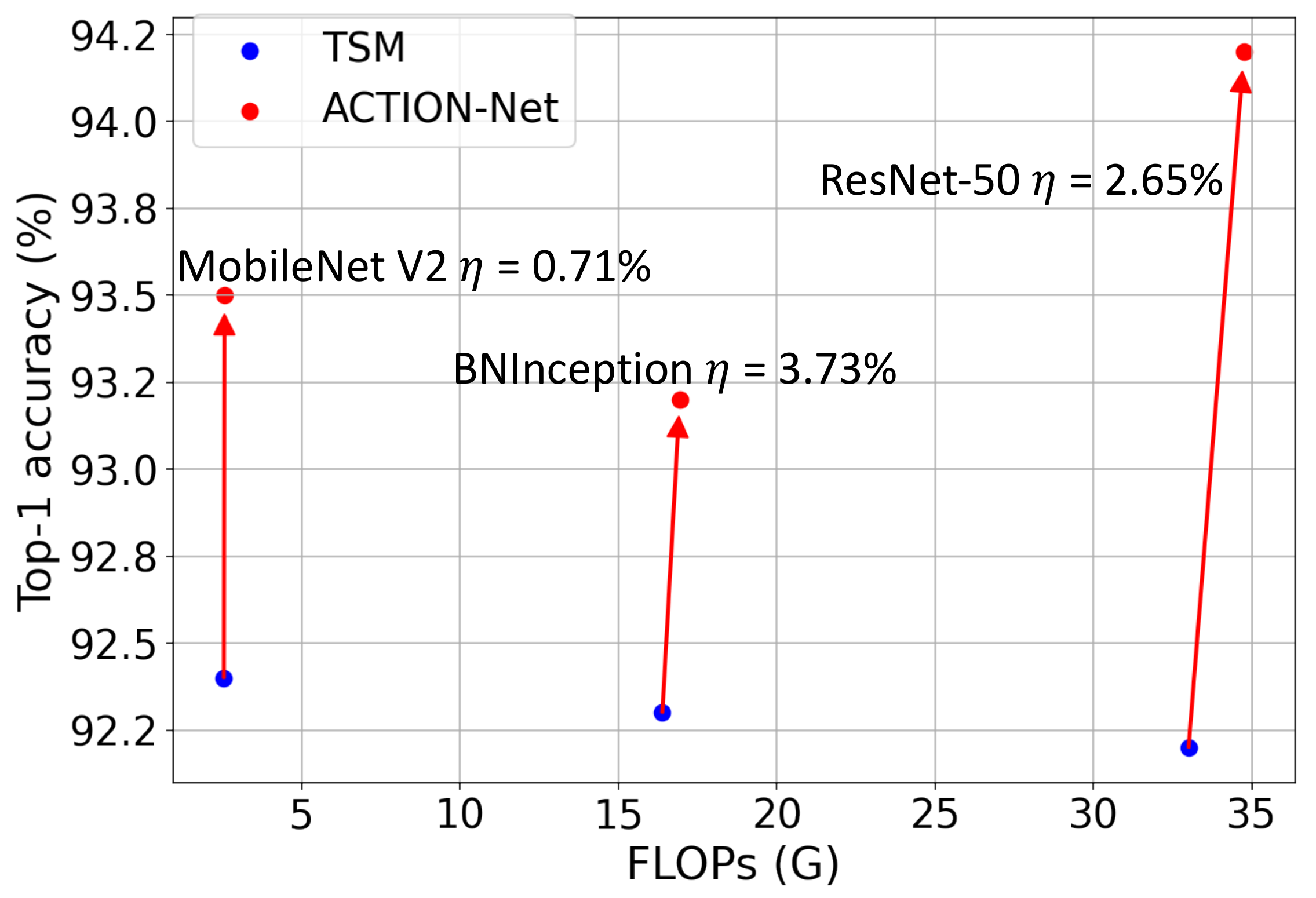}
        \label{fig:runtime_egogesture}    
    }
    \subfigure[Jester]{
        \centering
        \includegraphics[width=0.32\textwidth]{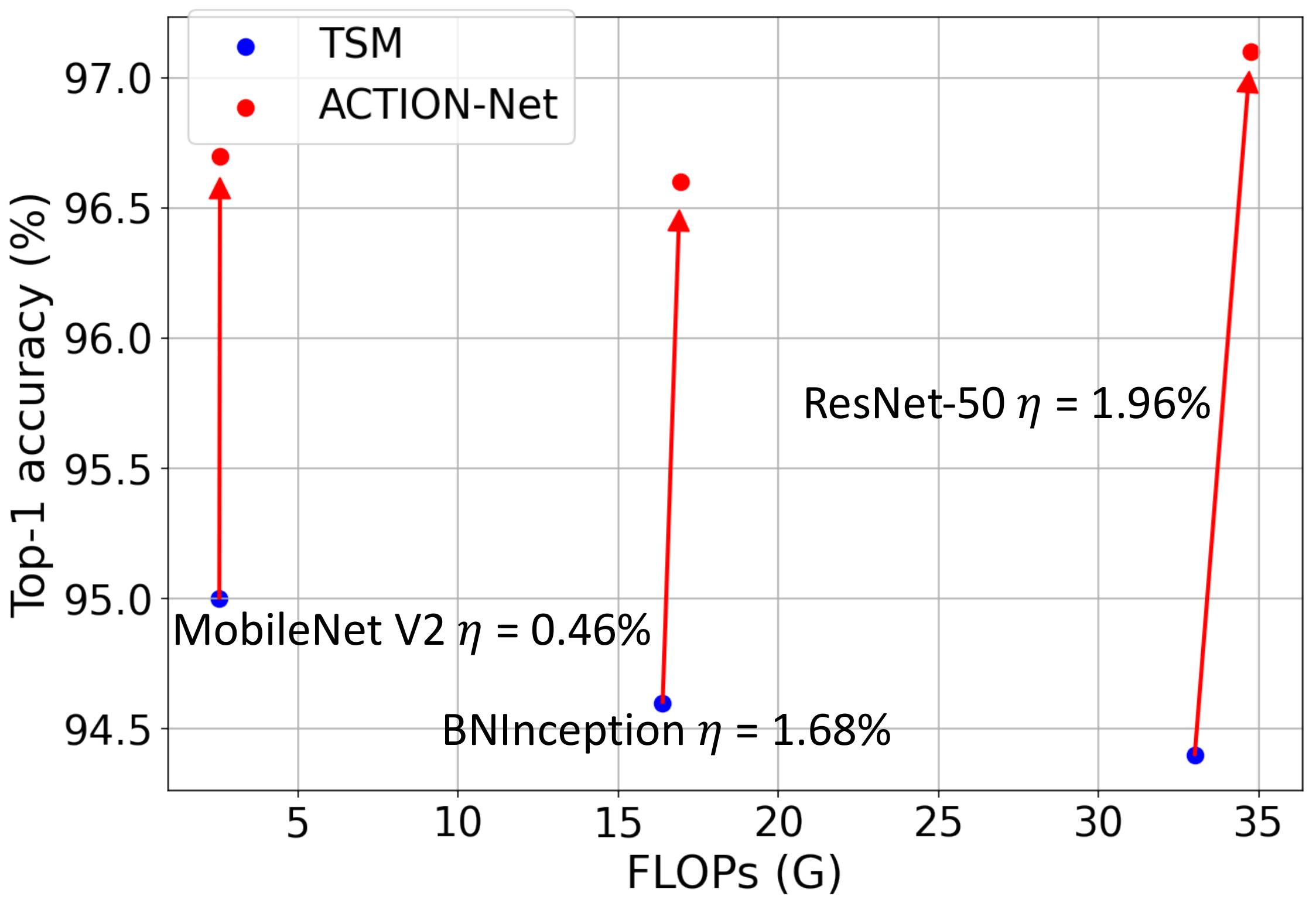}
        \label{fig:runtime_jester}    
    }
    \subfigure[Something-something V2]{
        \centering
        \includegraphics[width=0.32\textwidth]{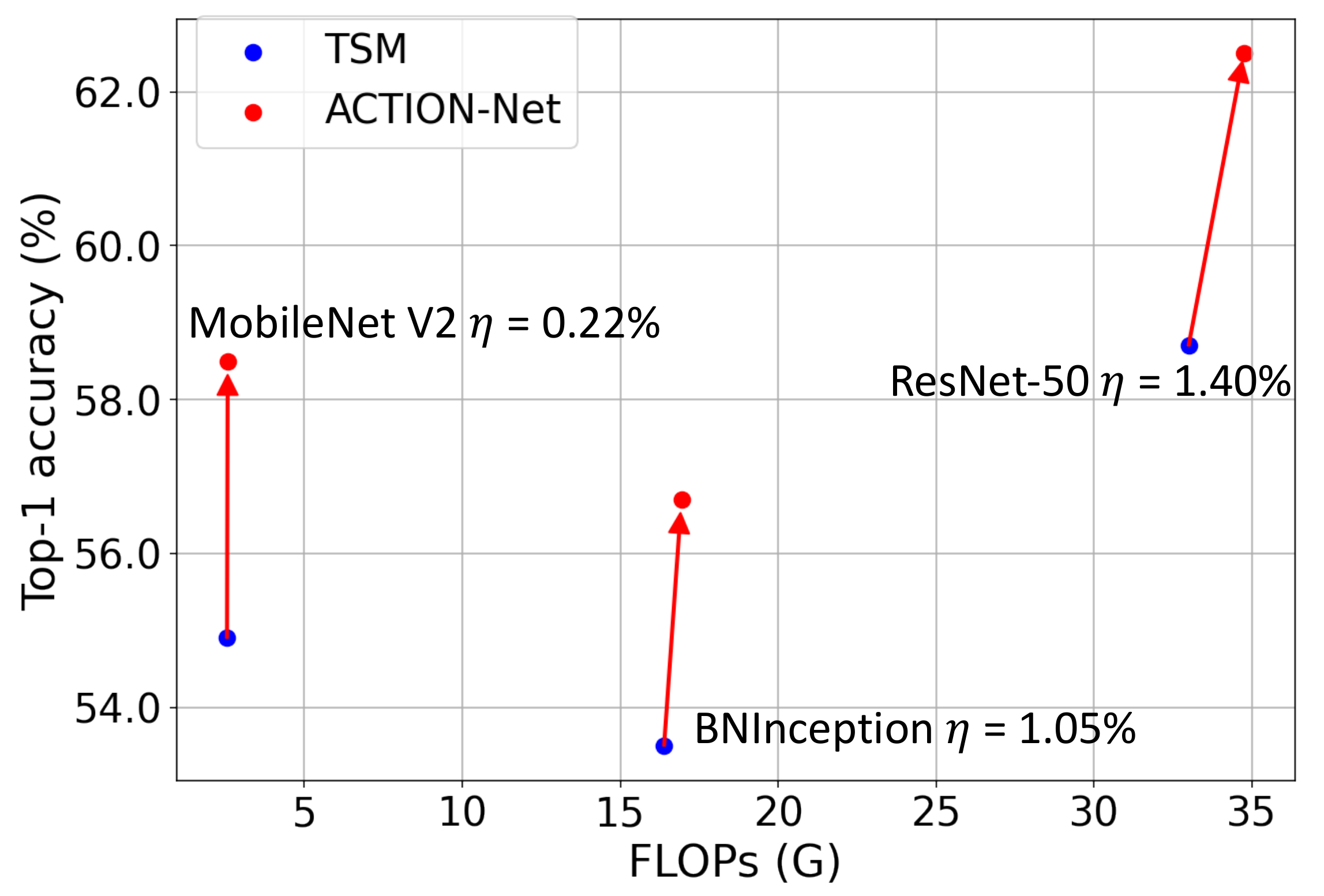}
        \label{fig:runtime_sthv2}    
    }
    \caption{Top-1 accuracy and FLOPs for ACTION-Net and TSM on three backbones i.e., ResNet-50, BNInception and MobileNet V2 using three datasets (from left to right: EgoGesture, Jester and Something-something V2). $\eta$ is calculated using equation~\eqref{eq:efficience} for each backbone on three datasets. EgoGesture: ResNet-50 $\eta=2.65\%$, BNInception $\eta=3.73\%$, MobileNet V2 $\eta=\textbf{0.71\%}$. Jester: ResNet-50 $\eta=1.96\%$, BNInception $\eta=1.68\%$, MobileNet V2 $\eta=\textbf{0.46\%}$. Something V2: ResNet-50 $\eta=1.40\%$, BNInception $\eta=1.05\%$, MobileNet V2 $\eta=\textbf{0.22\%}$.}
    \label{fig:runtime}
\end{figure*}
In this section, we investigate the design of our ACTION-Net with respect to (1) efficacy of each excitation and (2) impact of the number of ACTION modules in ACTION-Net regarding the ResNet-50 architecture. We carry out ablation experiments using 8 frames as the input on the EgoGesture dataset for inspecting these two aspects. 

\vspace{5pt}
\noindent
\textbf{Efficacy of Three Excitations.}\hspace{2pt} To validate the contribution of each excitation sub-module, we compare the performance for each individual module and the combination of all sub-modules (ACTION-Net) in Table~\ref{tab:three_paths}. We also provide visualization results in \textit{Supplementary Materials}. Results show that each excitation module is able to improve the performance for 2D CNN baselines provided by TSN and TSM with limited added computational cost. Concretely, STE and CE both add negligible extra computation compared to TSM by averaging channels globally and averaging spatial information globally yet they both provide useful information to the network. ME adds more computation and parameters to the network than the previous two yet it is acceptable. It captures temporal differences on the spatial domain among adjacent frames over the time and achieves better performance compared to STE and CE. When integrating all these three sub-modules to constitute the ACTION, it can be seen that the ACTION-Net achieves the highest accuracy and increases 2.1\% Top-1 accuracy together with increasing 5.3\% FLOPs. To better capture the relation between boosted performance and add-on computation, we define the efficiency formulated as
\begin{equation}
    \eta = \frac{\triangle\text{FLOPs}}{\triangle\text{Top-1}}
    \label{eq:efficience}
\end{equation}
where both $\triangle$FLOPs and $\triangle$Top-1 are in percent, $\eta$ is the efficiency that represents how many extra FLOPs \textit{in percent} are introduced with respect to increasing 1\% Top-1 accuracy (\textit{smaller indicates more efficient} apparently). Efficiency $\eta$ for STE, CE, ME and ACTION-Net is 0.18\%, 0.29\%, 2.83\% and 2.52\% respectively. It can be noticed that STE is the most efficient when taking $\eta$ into account. 

\vspace{5pt}
\noindent
\textbf{Impact of the Number of ACTION Blocks.}\hspace{2pt} The architecture of ResNet-50 can be divided into 6 stages i.e., $\text{conv}_{1}$, $\text{res}_{2}$, $\text{res}_{3}$, $\text{res}_{4}$, $\text{res}_{5}$ and FC. The ACTION module can be inserted into any residual stage from $\text{res}_{2}$ to $\text{res}_{5}$. We investigate the impact of the number of residual stages that contain the ACTION module as shown in Table~\ref{tab:block_num}. Results show that more stages including the ACTION module results in better performance, which indicates the efficacy for our proposed approach. 

\subsection{Analysis of Efficiency and Flexibility}\label{sec:Runtime}
Compared to recent 2D CNN approaches e.g., STM~\cite{jiang2019stm} and TEA~\cite{li2020tea}, our ACTION module enjoys a plug-and-play manner like TSM, which can be embedded to any 2D CNN. To validate the efficacy for our proposed approach, we compare ACTION-Net with TSM on three backbones i.e., ResNet-50, BNInception and MobileNet V2. We report FLOPs and Top-1 accuracy for ACTION-Net and TSM respectively. We also report $\eta$ calculated using equation~\ref{eq:efficience} when replacing TSM with ACTION, which indicates the penalization of computation when improving the accuracy. Table~\ref{tab:TSM_STCME} demonstrates recognition performance and computation for ACTION-Net employing three backbones. TSM is used as a baseline since TSM benefits good performance and zero extra introduced computational cost compared to TSN. It can be noticed that ACTION-Net outperforms TSM consistently regarding the accuracy for all three backbones yet with limited add-on computational cost. ResNet-50 is boosted most significantly regarding the performance and MobileNet V2 holds the least added FLOPs. Figure~\ref{fig:runtime} demonstrates the efficiency $\eta$ for ACTION-Net based on TSM on different backbones using three datasets. It can be noticed that MobileNet V2 achieves the lowest $\eta$ (the most efficient) while ResNet-50 achieves the highest $\eta$ (the least efficient) for all three datasets, which indicates that MobileNet V2 benefits mostly from the ACTION module regarding the efficiency.



\section{Conclusion}
We target at designing a module to be inserted to 2D CNN models for video action recognition and introduce a novel ACTION module that utilizes multipath excitation for spatio-temporal features, channel-wise features and motion features. The proposed module could be leveraged by any 2D CNN to build a new architecture ACTION-Net for video action recognition. We demonstrate efficacy and efficiency for ACTION-Net on three large-scale datasets. We show that ACTION-Net achieves consistently improvements compared to 2D CNN counterparts with limited extra computations introduced.

{\small
\bibliographystyle{ieee_fullname}
\bibliography{egbib}
}

\setcounter{figure}{0}
\setcounter{table}{0}
\setcounter{section}{0}
\clearpage
\renewcommand\thesection{\Alph{section}}
\section{Backbone Architecture in Experiments}\label{sup-sec:backbone}
In the main content of our paper, we evaluate efficiency and performance for our ACTION-Net on three different backbones i.e., ResNet-50, BNInception and MobileNet V2. We provide details that insert ACTION/TSM into each backbone in this section. To keep consistence, the number and the inserted position of TSM and ACTION are same. 

\vspace{5pt}
\noindent
\textbf{ResNet-50.}\hspace{2pt} We insert TSM/ACTION into each residual block i.e., from $\text{res}_2$ to $\text{res}_5$ at the start. It is summarized in Table~\ref{tab:sup_resnet_architecture}. There are 3, 4, 6, 3 TSM/ACTION modules that are inserted into $\text{res}_2$, $\text{res}_3$, $\text{res}_4$ and $\text{res}_5$ respectively. Therefore, ResNet-50 is equipped with 16 TSM/ACTION modules totally.

\vspace{5pt}
\noindent
\textbf{BNInception.}\hspace{2pt} Figure~\ref{fig:sup_BNInception} illustrates the details of BNInception used in our study. Similar to ResNet-50, we insert TSM/ACTION into each inception block at the starting point. In summary, there are 10 TSM/ACTION modules added into BNInception.

\begin{figure*}[!htbp]
    \centering
    \raisebox{-.5\height}{\includegraphics[width=0.25\textwidth]{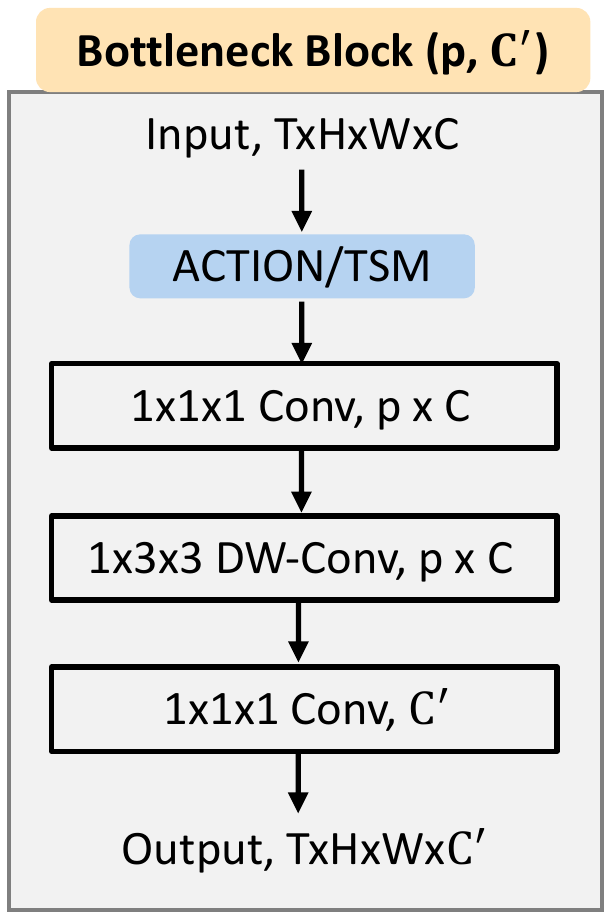}}
    \qquad
    \begin{tabular}{c|c|c}
         \toprule
         Stage& MobileNet V2 & Output size \\ \hline
         Input & ——& $T\times 224 \times 224$\\ \hline
         $\text{conv}_1$& $1\times 7\times 7, 32, \text{stride} 1, 2, 2$ & $T\times 112\times 112$ \\ \hline
         \multirow{2}{*}{$\text{Stage}_2$}& Bottleneck(1, 16) & \multirow{2}{*}{$T\times 56\times56$}\\ \cline{2-2}
                                          & Bottleneck(1, 16) $\times$ 2\\\hline
         $\text{Stage}_3$& Bottleneck(6, 25) $\times$ 3 & $T\times 28 \times 28$\\\hline
         \multirow{2}{*}{$\text{Stage}_4$}& Bottleneck(6, 64) $\times$ 4 & \multirow{2}{*}{$T\times 14\times14$}\\ \cline{2-2}
                                          & Bottleneck(6, 96) $\times$ 3 &\\ \hline
         \multirow{3}{*}{$\text{Stage}_5$}& Bottleneck(6, 160) $\times$ 3 & \multirow{3}{*}{$T\times 7\times7$}\\ \cline{2-2}
                                          & Bottleneck(6, 320)&\\ \cline{2-2}
                                          & $1\times 1\times 1, 1280, \text{stride} 1, 1, 1$&\\ \hline
        \multicolumn{2}{c|}{global average pool, FC, temporal average}& $CLS$\\
        \bottomrule
        
    \end{tabular}
    \captionlistentry[table]{Accuracy and model complexity on EgoGesture dataset with 8 frames as input. Three types of backbones are tested for TSM and our method.}
    \label{tab:sup_mobilenet_v2}
    \captionsetup{labelformat=andtable}
    \caption{\textit{Figure on the left}: Bottleneck block ($p$, $C^{'}$) with ACTION/TSM in MobileNet V2. We insert ACTION/TSM into the bottleneck block at the start. DW-Conv refers to depth-wise convolution~\cite{howard2017mobilenets}. \textit{Table on the right}: MobileNet-V2 backbone. Bottleneck blocks with ACTION/TSM illustrated in \textit{figure on the left} are applied to the backbone.~\cite{sandler2018mobilenetv2}.}
    \label{fig:sup_mobilenet_v2}
\end{figure*}

\vspace{5pt}
\noindent
\textbf{MobileNet V2.}\hspace{2pt} Figure~\ref{fig:sup_mobilenet_v2} and Table~\ref{eq:sub_module_output} summarize details of MobileNet V2 architecture and positions that insert TSM/ACTION. We insert TSM/ACTION into each bottleneck at the start. In order to keep consistent with adding TSM to MobileNet V2 in the original work~\cite{lin2019tsm}, we insert TSM/ACTION into two blocks in $\text{stage}_4$ and the first block in $\text{stage}_5$, which results in 10 TSM/ACTION modules have been added into MobileNet V2. 

\section{The Location of ACTION}\label{sup-sec:location}
\begin{figure*}[!htpb]
    \centering
    \includegraphics[width=1.\textwidth]{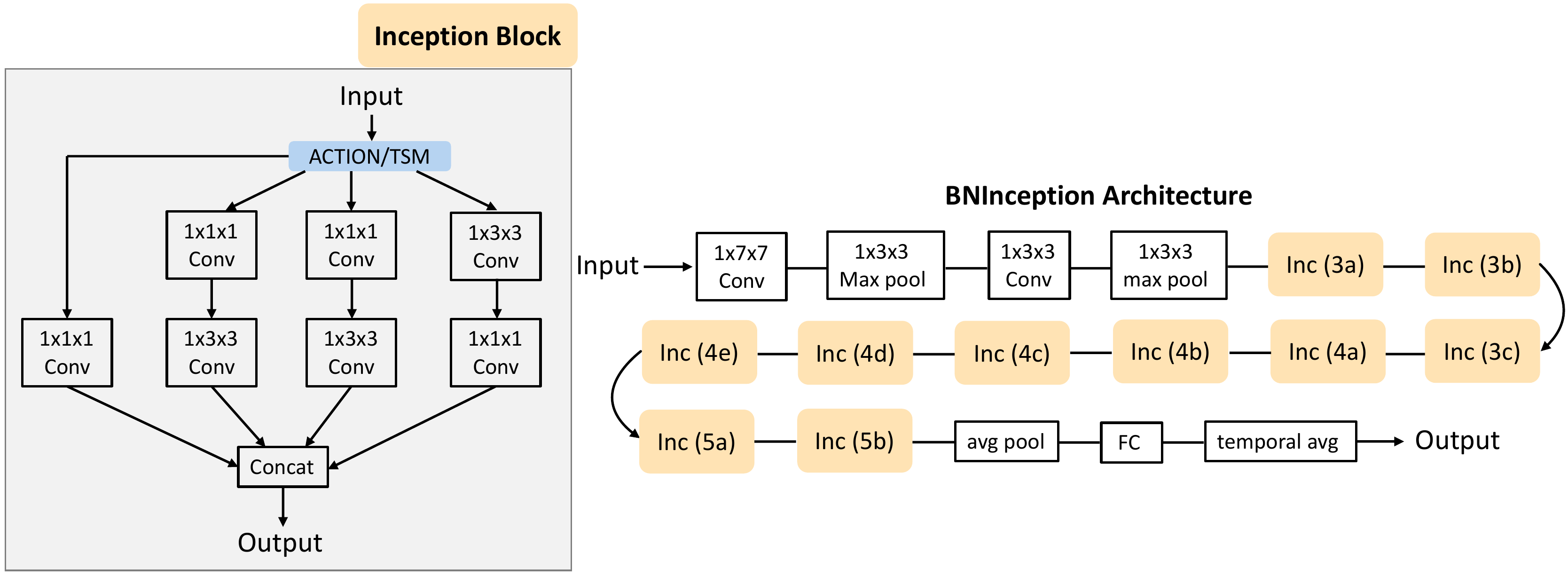}
    \caption{BNInception with ACTION and TSM used in this study. We insert ACTION/TSM into the start in each Inception block~\cite{ioffe2015batch}.}
    \label{fig:sup_BNInception}
\end{figure*}

As mentioned in the previous section, we insert ACTION at the beginning of each block for three backbones studied in this work. Here we provide ablation studies for different locations that insert ACTION to three backbones. It is worth nothing that four possible location for both ResNet-50 and MobileNet V2 but two possible locations for BNIncetion i.e., start and before the concatenate operation as seen in Fig.~\ref{fig:sup_BNInception}. It can be noticed that inserting ACTION at the beginning (Loc 1) is more effective compared to other locations for all three backbones.
\begin{table}[!htbp]
        \caption{Top-1 accuracy of different embedded locations on EgoGesture dataset using 8 segments. Loc 1 is the default used in the main paper.}
        \setlength{\tabcolsep}{1mm}{
        \begin{tabular}{ccccc}
            \toprule
             {Model}& Loc 1 (default)& Loc 2 & Loc 3 & Loc 4 \\ 
             \midrule
             {ResNet-50} &  \textbf{94.2}& 94.0& 93.8& 94.0\\ 
             {BNInception} &  \textbf{93.2}& 92.6& NA& NA\\ 
             {MobileNet V2} &  \textbf{93.5}& 93.1& 93.1& 93.3\\ 
             \bottomrule
        \end{tabular}
        \label{tab:location}
        }
\end{table}

\begin{table*}[!htpb]
    \centering
    \caption{ResNet-50 backbone with TSN, TSM and ACTION used in this work.}
    \begin{tabular}{c|c|c|c|c}
        \toprule
        Stage & TSN & TSM & ACTION-Net& Output size \\ \hline
        Input & \multicolumn{3}{c|}{——} & $T\times 224 \times 224$ \\ \hline
        $\text{conv}_1$ & \multicolumn{3}{c|}{$1\times7\times 7$, 64, stride 1, 2, 2} & $T\times 112 \times 112$ \\ \hline
        $\text{pool}_1$ & \multicolumn{3}{c|}{$1\times3\times 3$, max, stride 1, 2, 2} & $T\times 56 \times 56$ \\ \hline
        $\text{res}_2$& $\Bigg[\begin{array}{c}
                                1\times1\times1, 64\\
                                1\times3\times3, 64\\
                                1\times1\times1, 256\\
                              \end{array}  
                        \Bigg]\times3$ &
                        $\Bigg[\begin{array}{c}
                                \text{TSM}\\
                                1\times1\times1, 64\\
                                1\times3\times3, 64\\
                                1\times1\times1, 256\\
                              \end{array}  
                        \Bigg]\times3$ &
                        $\Bigg[\begin{array}{c}
                                \text{ACTION}\\
                                1\times1\times1, 64\\
                                1\times3\times3, 64\\
                                1\times1\times1, 256\\
                              \end{array}  
                        \Bigg]\times3$ &
                        $T\times 56 \times 56$ \\ \hline
        $\text{res}_3$& $\Bigg[\begin{array}{c}
                                1\times1\times1, 128\\
                                1\times3\times3, 128\\
                                1\times1\times1, 512\\
                              \end{array}  
                        \Bigg]\times4$ &
                        $\Bigg[\begin{array}{c}
                                \text{TSM}\\
                                1\times1\times1, 128\\
                                1\times3\times3, 128\\
                                1\times1\times1, 512\\
                              \end{array}  
                        \Bigg]\times4$ &
                        $\Bigg[\begin{array}{c}
                                \text{ACTION}\\
                                1\times1\times1, 128\\
                                1\times3\times3, 128\\
                                1\times1\times1, 512\\
                              \end{array}  
                        \Bigg]\times4$ &
                        $T\times 28 \times 28$ \\ \hline
        $\text{res}_4$& $\Bigg[\begin{array}{c}
                                1\times1\times1, 256\\
                                1\times3\times3, 256\\
                                1\times1\times1, 512\\
                              \end{array}  
                        \Bigg]\times6$ &
                        $\Bigg[\begin{array}{c}
                                \text{TSM}\\
                                1\times1\times1, 256\\
                                1\times3\times3, 256\\
                                1\times1\times1, 1024\\
                              \end{array}  
                        \Bigg]\times6$ &
                        $\Bigg[\begin{array}{c}
                                \text{ACTION}\\
                                1\times1\times1, 256\\
                                1\times3\times3, 256\\
                                1\times1\times1, 1024\\
                              \end{array}  
                        \Bigg]\times6$ &
                        $T\times 14 \times 14$ \\ \hline
        $\text{res}_5$& $\Bigg[\begin{array}{c}
                                1\times1\times1, 512\\
                                1\times3\times3, 512\\
                                1\times1\times1, 2048\\
                              \end{array}  
                        \Bigg]\times3$ &
                        $\Bigg[\begin{array}{c}
                                \text{TSM}\\
                                1\times1\times1, 512\\
                                1\times3\times3, 512\\
                                1\times1\times1, 2048\\
                              \end{array}  
                        \Bigg]\times3$ &
                        $\Bigg[\begin{array}{c}
                                \text{ACTION}\\
                                1\times1\times1, 512\\
                                1\times3\times3, 512\\
                                1\times1\times1, 2048\\
                              \end{array}  
                        \Bigg]\times3$ &
                        $T\times 7 \times 7$ \\ \hline
       \multicolumn{4}{c|}{global average pool, FC}& $T\times CLS$\\ \hline
       \multicolumn{4}{c|}{temporal average}& $CLS$\\
       \bottomrule
    \end{tabular}
    \label{tab:sup_resnet_architecture}
\end{table*}

\section{Visualization Results}
\begin{figure*}[!htpb]
    \centering
    \includegraphics[width=1.\textwidth]{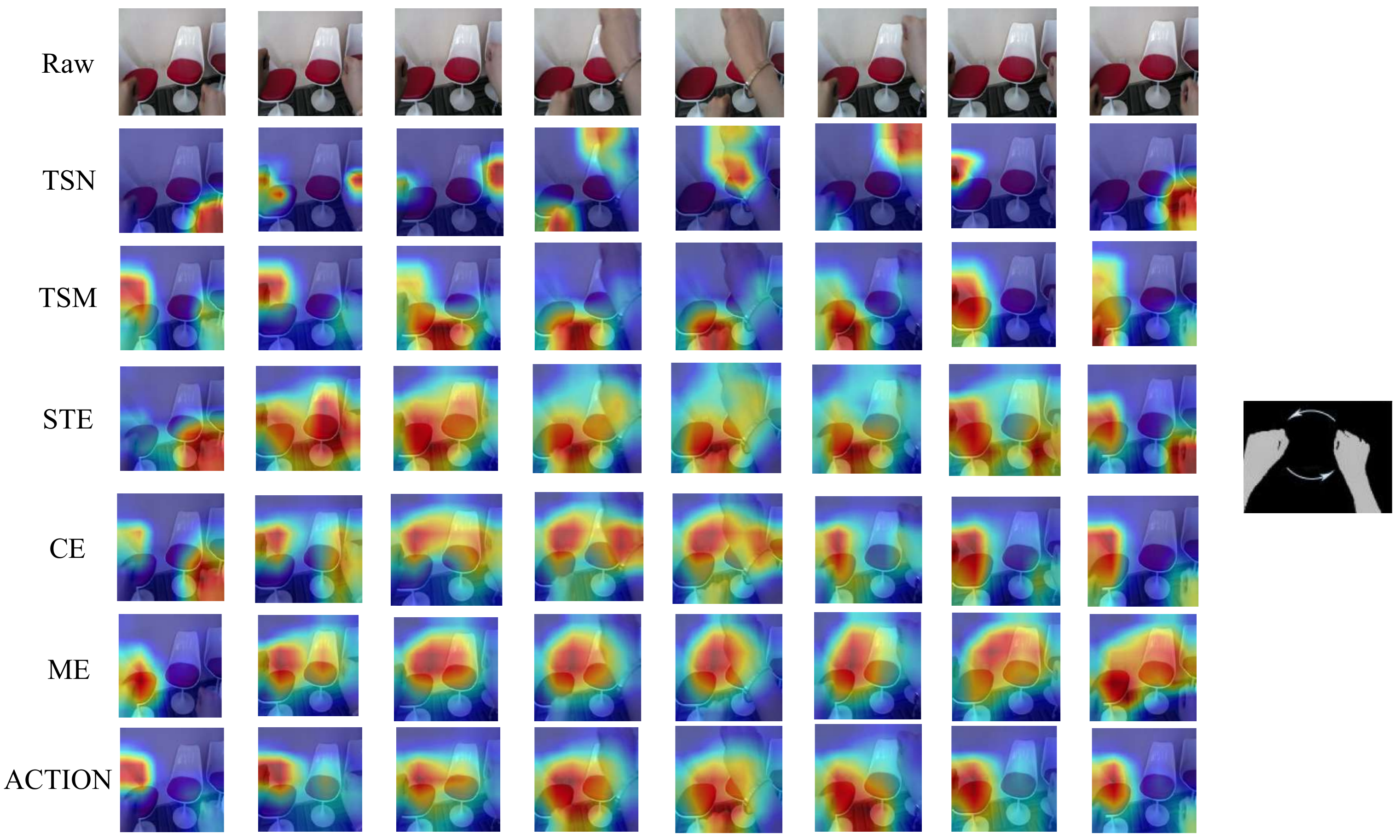}
    \caption{\textit{`Rotate fists counterclockwise'}}
    \label{fig:supp_heatmap_10}
\end{figure*}

\begin{figure*}[!htpb]
    \centering
    \includegraphics[width=1.\textwidth]{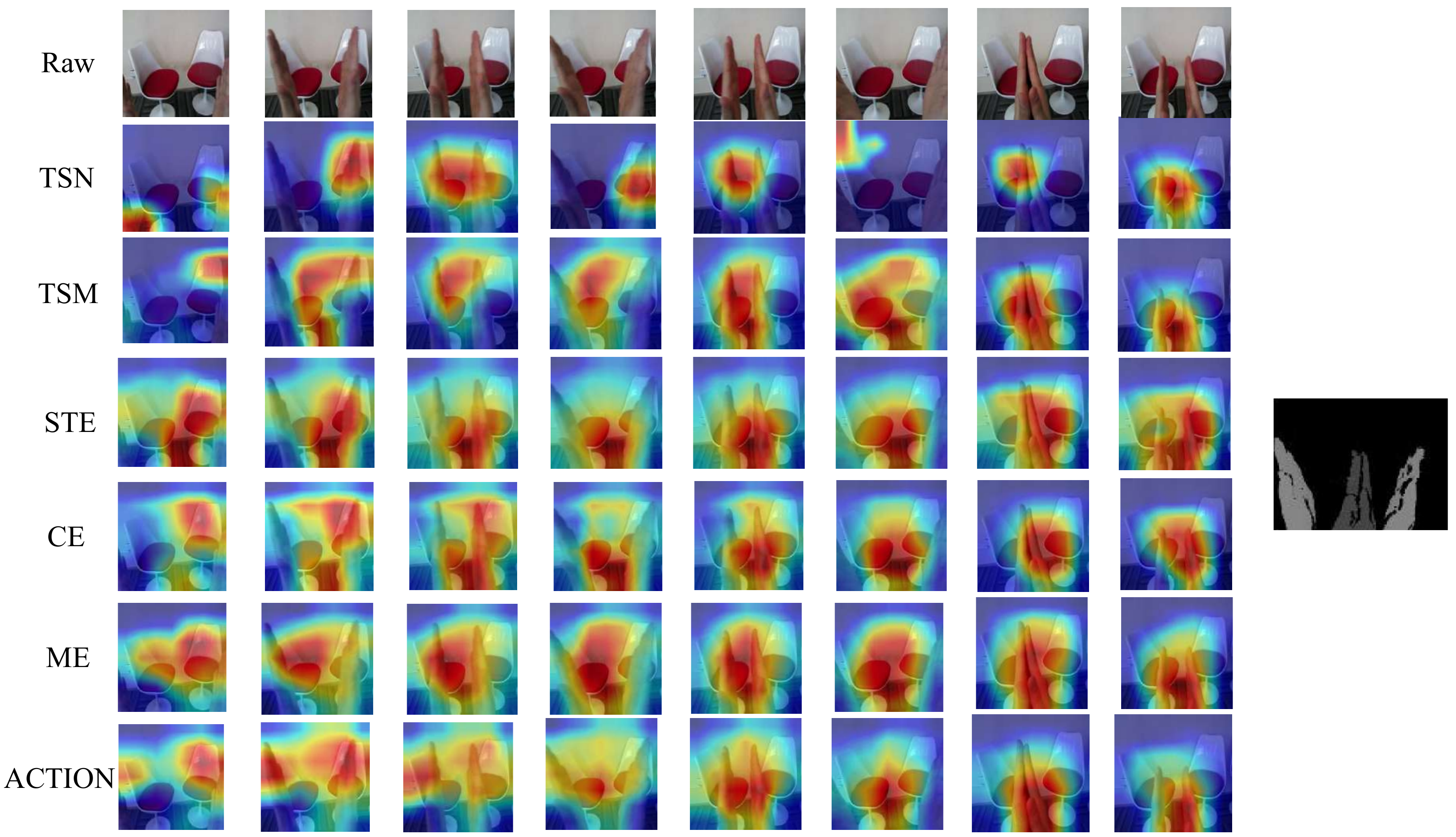}
    \caption{\textit{`Applaud'}}
    \label{fig:supp_heatmap_51}
\end{figure*}

\begin{figure*}[!htpb]
    \centering
    \includegraphics[width=1.\textwidth]{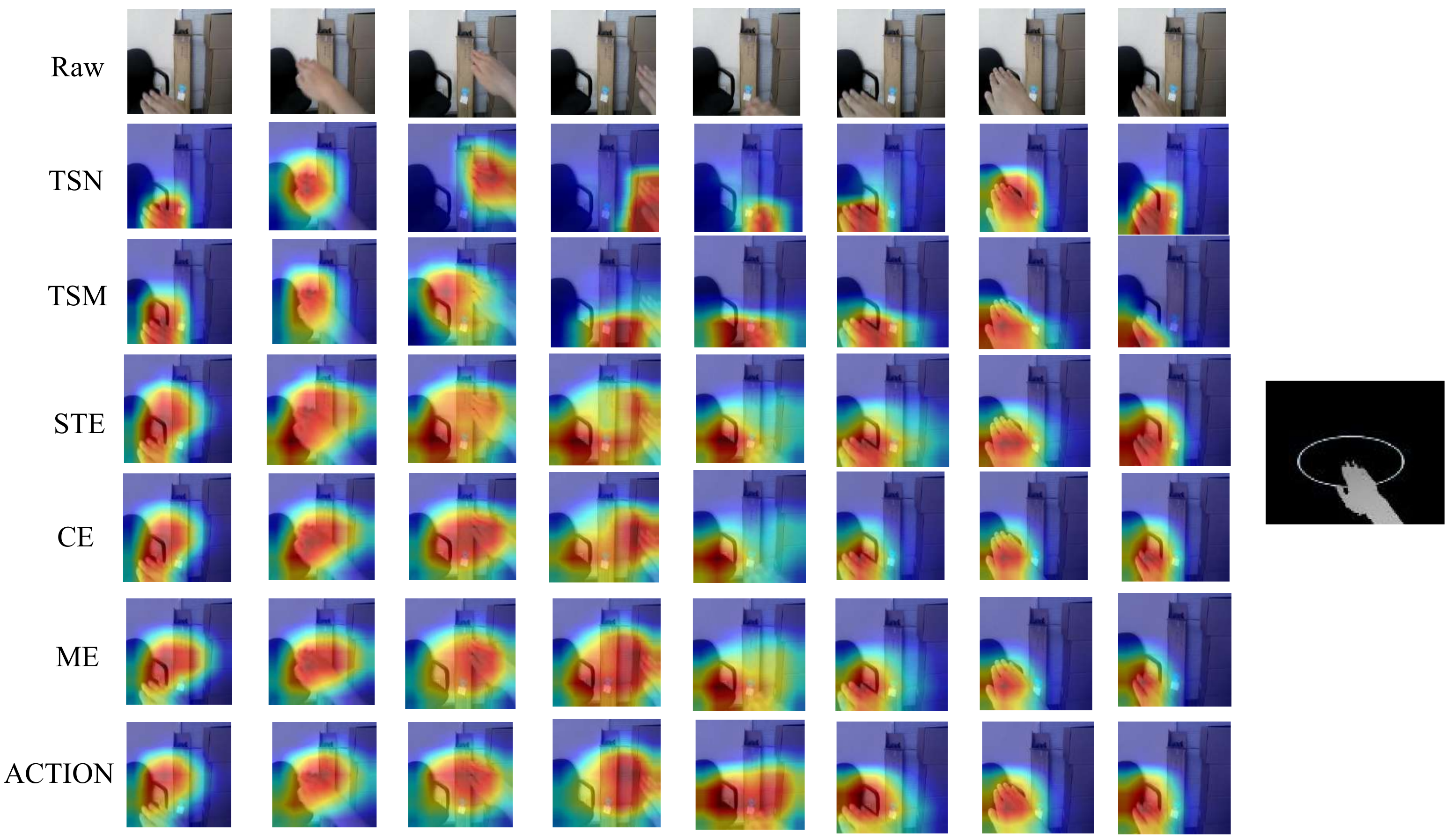}
    \caption{\textit{`Draw circle with hand in horizontal surface'}}
    \label{fig:supp_heatmap_72}
\end{figure*}

Visualization for three actions \textit{`Rotate fists counterclockwise'}, \textit{`Applaud'} and \textit{`Draw circle with hand in horizontal surface'} using two baselines (i.e., TSN and TSM), three excitation sub-modules and our proposed ACTION-Net is shown in Fig.~\ref{fig:supp_heatmap_10}, Fig.~\ref{fig:supp_heatmap_51} and Fig.~\ref{fig:supp_heatmap_72}. The first row is the presented video sequence and rest of rows are CAM obtained by each approach. It can be noticed that both TSN and TSM can only recognize objects but are unable to produce CAM for the movement smoothly. Compared to TSN and TSM, it can be noticed that our proposed ACTION and each three sub-module are all able to extract meaningful temporal information from a presented video sequence by addressing smooth CAM for the action movement. Although STE and CE more focus on spatial modeling since the temporal modeling ability is limited in these two modules i.e., one 3D convolutional layer with size $3\times3\times3$ and one 1D convolutional layer with size $3$, they are able to produce smooth CAM to some extent, which are much more convincing than TSM and TSN. From the visualization for ME, it can be noticed that ME is able to produce the most smooth CAM for the action movement between adjacent frames. However, spatial information for objects in video is somewhat limited e.g., it is hard to figure out one hand or two hands in the presented video for Fig.~\ref{fig:supp_heatmap_10} and Fig.~\ref{fig:supp_heatmap_51}. Our proposed ACTION-Net, which integrates three excitation above, is able to not only recognize objects (first two frames) but also address action movements (middle frames), which takes advantages from each excitation sub-module. 

\end{document}